\documentclass[letterpaper, 10 pt, journal, twoside]{IEEEtran}

\usepackage[linesnumbered,ruled,vlined]{algorithm2e}

\usepackage{color}
\definecolor{darkred}{RGB}{205,38,38}

\usepackage{caption}
\usepackage{subfig}
\usepackage{float} 
\let\labelindent\relax 
\usepackage{enumitem}
\usepackage{booktabs} 
\usepackage{multirow}
\usepackage{cite}
\usepackage{array}
\newcolumntype{L}[1]{>{\raggedright\let\newline\\\arraybackslash\hspace{0pt}}m{#1}}
\newcolumntype{C}[1]{>{\centering\let\newline\\\arraybackslash\hspace{0pt}}m{#1}}
\newcolumntype{R}[1]{>{\raggedleft\let\newline\\\arraybackslash\hspace{0pt}}m{#1}}
\usepackage{threeparttable}
\usepackage{makecell}
\usepackage{balance}
\usepackage{amsmath}
\usepackage{amssymb,bm}
\usepackage{amsfonts}
\usepackage{enumerate}
\usepackage{graphicx}
\usepackage{pifont}

\usepackage{enumitem}
\usepackage{hyperref}
\hypersetup{
	colorlinks=true,
	linkcolor=darkred, %color of \ref
	filecolor=darkred,      
	urlcolor=red,
	citecolor=blue, %colore of \cite
}

\title{CollaBot: Vision-Language Guided Simultaneous Collaborative Manipulation of Large Objects}

\author{
	Kun Song$^{1}$, Shentao Ma$^2$, Gaoming Chen$^2$, Ninglong Jin$^2$, Guangbao Zhao$^2$, \\Mingyu Ding$^{3*}$, Zhenhua Xiong$^{2*}$, and Jia Pan$^{1*}$  
	\thanks{$^1$K. Song, and J. Pan are with the Department of Computer Science, The University of Hong Kong, Hong Kong SAR, China. (e-mail: kunsonghku@connect.hku.hk, jpan@cs.hku.hk).
		}
	\thanks{$^2$S. Ma, G. Chen, N. Jin, G. Zhao, and Z. Xiong are with the School of Mechanical Engineering, Shanghai Jiao Tong University, Shanghai, China. (e-mail: \{shentao\_ma, cgm1015, ninglong\_jin, zhaoguangbao, mexiong\}@sjtu.edu.cn)
		}
	\thanks{$^3$M. Ding is with the Department of Computer Science at at UNC-Chapel Hill, Chapel Hill, USA. (e-mail: md@cs.unc.edu)
		}
	\thanks{
			$^*$Corresponding author.
		}
}

\begin{document}

\maketitle
\thispagestyle{empty}
\pagestyle{empty}

\begin{abstract}
	One central goal of robotics is to enable robots to interact with the physical world. 
	Traditional manipulation studies primarily focus on single robots and relatively small objects. 
	However, factory and domestic environments often require large-object manipulation, such as moving tables, where multiple robots must work collaboratively.
	Existing studies still lack a generalizable framework that can handle diverse objects, tasks, and robot team sizes. 
	In this work, we propose CollaBot, a generalist framework for simultaneous collaborative manipulation.
	First, we use SEEM for scene segmentation and target-object extraction. 
	Then, we propose a collaborative grasping framework that decomposes the task into local grasp pose generation and global coordination. 
	Finally, we design a two-stage planning module to generate collision-free trajectories for task execution.
	Experimental results across different settings with varying objects, tasks, and numbers of robots indicate that our framework achieves a 72\% success rate. 
	This marks a substantial improvement over behavior cloning-based methods, validating the advantages of the proposed framework in complex multi-robot cooperative tasks.
	Real-world experiments further demonstrate the feasibility of our method in practical applications.
\end{abstract}

\section{Introduction}

Multi-robot collaborative manipulation uses multiple robots to perform tasks that exceed the capabilities of a single robot \cite{lai2025roboballet,mandi2024roco,qin2025robofactory}. 
By extending manipulation ability from small objects to large items, this approach offers significant potential for applications in factory workshops, logistics, and domestic environments.
Collaborative manipulation tasks can generally be categorized into two types: sequential processes\cite{mandi2024roco,zhang2024lamma}, where robots perform operations one after another, such as in an assembly line where one robot completes a task and then passes the object to the next robot, which can also be called hand-over cooperation\cite{shome2019anytime}; 
and simultaneous collaborative manipulation (SCM)\cite{feng2020overview,liu2025planning,qin2025robofactory}, where multiple robots interact with the same object at the same time, such as jointly transporting a large item. 
% For sequential processes, a heterogeneous robotic system can be employed to improve the efficiency\cite{chen2024textbf}. 
For SCM, the system's overall manipulation capability can be significantly enhanced by enabling the manipulation of large objects, which presents broad application prospects \cite{feng2020overview, zhou2022topp}.

\begin{figure}[!t]
	\centering   
	\includegraphics[width=3.3in]{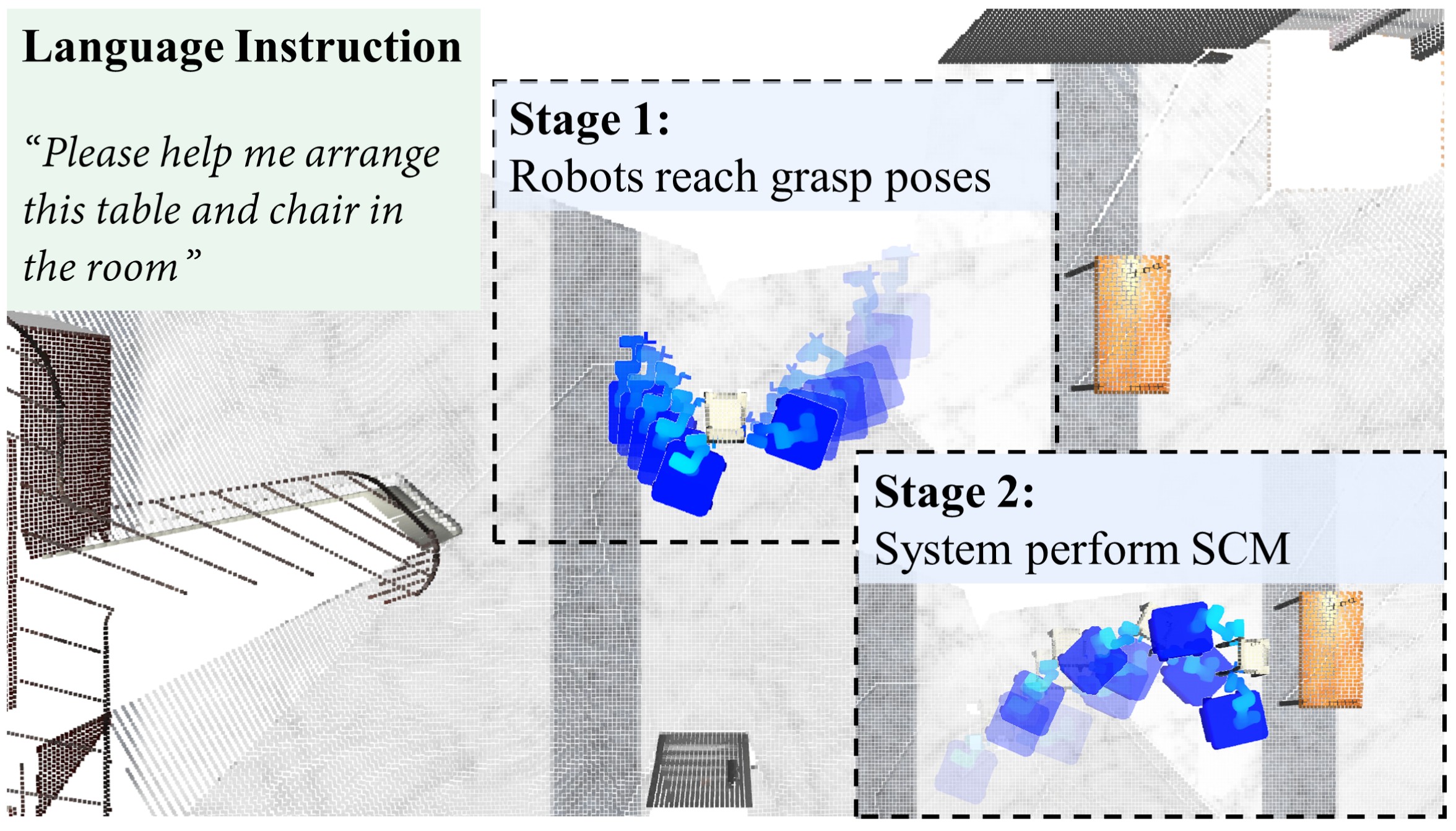}
	\caption{
		An example of the proposed framework.
		Given a language instruction, the system uses VLMs to generate motion constraints.
		Then, grasp poses are generated for large object manipulation.
		Finally, a two-staged motion planning algorithm is performed to accomplish the task.}
	\label{fig:fig1}
\end{figure}

However, SCM is relatively under-explored. 
Previous studies exhibit several limitations. 
First, many methods restrict the number of robots in the team and are only applicable to two robots\cite{gao2024coohoi,zhou2022topp,zhai20222,karim2025dg16m}. 
Second, most existing work primarily focuses on motion planning \cite{qin2025robofactory,liu2025planning,matsuura2023development}, while paying limited attention to task and grasp planning throughout the manipulation process, especially for large objects, which limits generalization and makes these solutions difficult to deploy in real-world environments.

Due to the inherently higher degrees of freedom in multi-robot systems, it is challenging to collect sufficient data to enable collaborative manipulation through methods like imitation learning \cite{xie2020deep,brohan2022rt,brohan2023rt}. 
More importantly, during collaborative transportation, it is essential to maintain fixed relative poses between the end-effectors of multiple robots, which introduces significant challenges for data collection based on teleoperation\cite{aldaco2024aloha}. 
Moreover, the collected data is often difficult to generalize to a system with more robots.
In recent years, with the advancement of foundation models, an increasing number of studies have leveraged the prior knowledge embedded in these models to achieve single-robot manipulation\cite{huang2024copa,huang2023voxposer,huang2024rekep,intelligence2026pi,wu2024selp}. 
However, due to the internal complexity of multi-robot systems, how to effectively use foundation models for SCM remains an open problem.

Therefore, this paper investigates how to leverage the guidance of foundation models to enable collaborative manipulation tasks under arbitrary instruction inputs. 
We assume the presence of a given robot team, such as multiple mobile manipulators (MMMs), randomly distributed in the environment. 
The primary objective is to autonomously determine the motion of each individual robot based on the language input in order to accomplish the task.

SCM tasks with multi-robot systems introduce three new challenges.
First, the workspace extends from a fixed-size environment that can be captured by a single camera to a larger, open environment, requiring broader scene understanding. 
Second, the system must provide a suitable manipulation framework that can adapt to different object types and size. 
Third, the overall system must integrate scene understanding, grasp generation, and motion planning into a coherent pipeline that works reliably for SCM tasks.

Therefore, we propose CollaBot, a framework for \underline{colla}borative manipulation using multi-ro\underline{bot} systems.
First, based on the input instruction, we use a Large Language Model (LLM) for language understanding to identify the target object for manipulation, and employ SEEM\cite{zou2023segment} to extract the corresponding object point cloud. 
In addition, we adopt a prompt-based in-context learning approach \cite{arenas2024prompt} to generate motion constraints for the object during manipulation.
Second, inspired by how humans make decisions when moving a large object, the grasp planning is divided into local and global stages. 
We construct a dataset to evaluate local grasp stability for large objects using ShapeNet \cite{chang2015shapenet,zhai20222}, and propose LoGNet (Large Object Grasp Net) for local grasp pose generation. 
Global grasp planning is achieved through the guidance of a Vision-Language Model (VLM).
Third, we propose a two-stage planning framework to generate feasible trajectories for task execution.
An example of SCM using robots can be found in Fig. \ref{fig:fig1}.

Our main contributions are as follows:
\begin{itemize}[leftmargin=*]
	\item We propose CollaBot, a general framework for SCM of large objects with multiple robots.
	\item We design three key components for SCM: VLM-guided scene understanding and constraint inference, collaborative grasp pose generation, and two-stage motion planning.
	\item We conduct extensive experiments in both simulation and the real world to validate the effectiveness and deployability of the proposed method.
\end{itemize}

\section{Related Work on SCM}
SCM has been studied over the past three decades\cite{rus1995moving}, and most research efforts focus on developing planning algorithms for SCM\cite{shorinwa2020scalable,liu2025planning,zhang2025reactive}.
For mobile manipulators, the degrees of freedom (DoF) typically exceed six, making their motion planning inherently a high-dimensional redundancy problem.
Furthermore, with multiple mobile manipulators, planning becomes even more challenging due to the significantly increased overall system DoFs.
In \cite{shorinwa2020scalable}, a distributed motion planning algorithm is proposed to enable large object manipulation.
Similarly, in \cite{liu2025planning}, a planning framework for MMMs that combines motion and regrasping planning is proposed with pre-defined grasp poses.
However, these methods often rely on several assumptions, such as complete knowledge of the objects, a fixed number of robots, and predefined grasp poses.
This implies that these methods have limited generalizability in the real world.
In \cite{lai2025roboballet,qin2025robofactory}, learning-based methods have been used for multi-robot collision-free trajectory generation.
These approaches focus primarily on tasks with minimal impact on the object, such as painting, and are therefore difficult to extend to laTrge-scale manipulation of large objects.

Multi-robot collaborative manipulation shares similarities with dual-arm manipulation, as two arms can be considered as two robots.
%For imitation learning, data collection is often challenging. 
Since dual-arm robotic systems closely correspond to human arms, data collection for such systems is relatively more feasible. 
In \cite{liu2024rdt}, RDT-1B was proposed based on Diffusion Transformers and imitation learning.
In contrast, collecting data for MMMs via teleoperation is significantly more difficult. 
As a result, few studies have focused on using imitation learning to enable manipulation tasks based on MMMs.
An alternative and promising direction lies in leveraging the knowledge embedded in pre-trained foundation models to enable task execution based on natural language instructions like \cite{mandi2024roco,huang2024copa,huang2024rekep}.

%In this work, we present a framework for large-object manipulation using MMMs under a minimal subtask decomposition input. 
%Leveraging foundation models, we address several challenging aspects of multi-robot manipulation, including scene understanding, motion constraint generation, and grasp pose synthesis.

\begin{figure*}[!t]
	\centering   
	\includegraphics[width=7in]{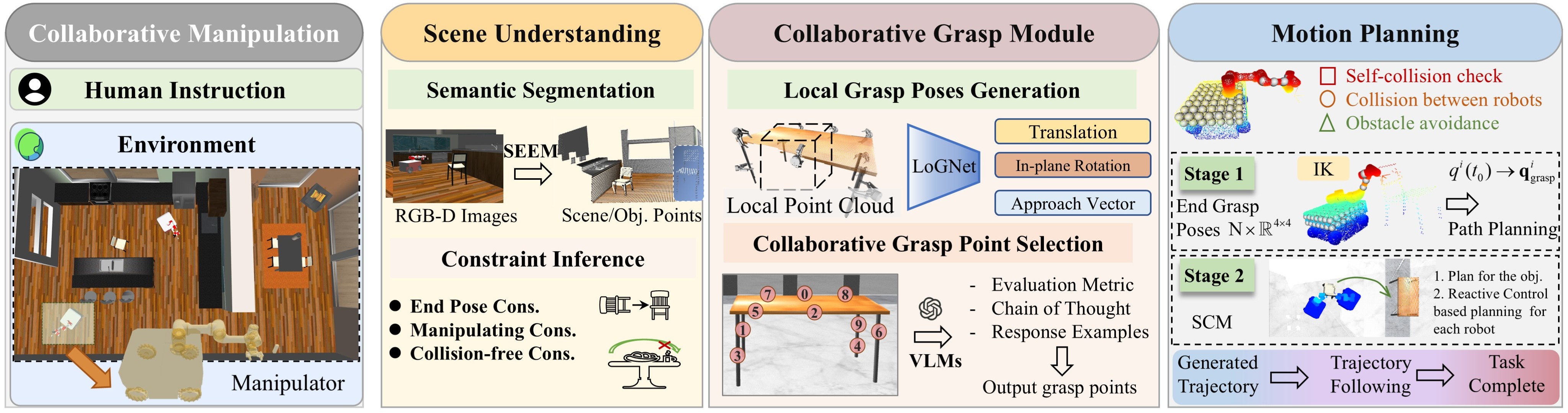}
	\caption{
		The overall framework of the proposed method.
		CollaBot is divided into three parts.
		First, scene understanding is performed to generate a representation of the environment, including spatial point clouds and task-relevant constraints. 
		Then, a two-step collaborative grasping module is applied.
		Finally, the motion planning algorithms generate trajectories for all robots to accomplish the task.
	}
	\label{fig:total}
\end{figure*}

\section{Problem Formulation}
%Assume there are $N \in \mathbb{Z}^+$ robots in the environment, each represented by $\mathcal{R}^i$. 
%Each robot is equipped with an end-effector, whose pose at time $t$ is denoted by $\mathbf{e}(t) \in SE(3)$. 
%The configuration space pose of $\mathcal{R}^i$ is represented by $\mathbf{q}^i\in \mathbb{R}^k$, where $k$ is the DoF for each robot. 

Assume there are $N \in \mathbb{Z}^+$ robots in the environment, each denoted by $\mathcal{R}^i$. 
The configuration of robot $\mathcal{R}^i$ is represented by the augmented vector $\mathbf{q}^i = [(\mathbf{q}_a^i)^\top, q_g^i]^\top \in \mathbb{R}^{k+1}$, where $\mathbf{q}_a^i \in \mathbb{R}^k$ corresponds to the $k$ robot DoFs, including the mobile base and manipulator, and $q_g^i \in \mathbb{R}$ characterizes the actuation state (e.g., opening or closing) of the end-effector. 
The pose of the end-effector at time $t$ is denoted by $\mathbf{e}^i(t) \in SE(3)$, which is determined by $\mathbf{q}_a^i(t)$ through the forward kinematics, while $q_g^i$ controls the gripper state.

Similar to fixed single-arm manipulation\cite{huang2023voxposer,huang2024rekep,intelligence2026pi}, collaborative manipulation can also be decomposed into a sequence of subtasks, each involving a specific operation on a particular object. 
For example, a subtask may involve moving a table from its initial state to a desired final state. 
Therefore, for a free-form instruction, we assume there exists a task planner that can decompose this instruction into subtasks.
A specific subtask can be represented by $\mathcal{T}=[O, \mathbf{p}_{\text{init}}, \mathbf{p}_{\text{end}}]$. 
Here, $O$ is the object to be manipulated.
$\mathbf{p}_{\text{init}}$ and $\mathbf{p}_{\text{end}}$ belong to $SE(3)$ and denote the initial and final poses of $O$, respectively.
Note that all poses are defined in a global world frame, like the initial base frame of the first robot.

We assume these $N$ robots are deployed for the task.
We divide the SCM task into two main stages. 
First, the robots move to the generated grasp poses. 
Then, all robots perform collaborative manipulation under end-effector pose constraints. 
Achieving $\mathcal{T}$ is equivalent to solving a motion planning problem defined by Eq. \ref{equ:goal}

%\begin{equation}
%	\label{equ:goal}
%	\begin{split}
	%		&\min \ |I_u(\mathcal{T})|\\
	%		\text{s.t.}&\ \left\{
	%		\begin{array}{lc}
		%			\mathbf{q}^i(t_1)= \mathbf{q}_{\text{grasp}}^i,\forall i\in I_u(\mathcal{T})\\
		%			\mathbf{p}(t_1)=\mathbf{p}_{\text{init}},\mathbf{p}(t_e)=\mathbf{p}_{\text{end}}\\
		%			\dot{\mathbf{T}}(\mathbf{e}_i,\mathbf{e}_j,t)=\mathbf{0},\forall i,j \in I_u(\mathcal{T}),t\in [t_1,t_e]\\
		%			\mathbf{G}(\mathbf{p}(t))\leq \mathbf{0},t\in [t_1,t_e].\\
		%		\end{array}\right.
	%	\end{split}
%\end{equation}
\begin{equation}
	\label{equ:goal}
	\begin{split}
		&\text{Find} \ \mathbf{q}^i(t), \mathbf{q}_{\text{grasp}}^i\\
		\text{s.t.}&\ \left\{
		\begin{array}{lc}
			\mathbf{q}^i(t_1)= \mathbf{q}_{\text{grasp}}^i,\forall i\in \{1, \dots, N\}\\
			\mathbf{p}(t_1)=\mathbf{p}_{\text{init}},\mathbf{p}(t_e)=\mathbf{p}_{\text{end}}\\
			\dot{\mathbf{T}}(\mathbf{e}_i,\mathbf{e}_j,t)=\mathbf{0},\forall i,j \in \{1, \dots, N\},t\in [t_1,t_e]\\
			\mathbf{G}(\mathbf{p}(t))\leq \mathbf{0},t\in [t_1,t_e].\\
		\end{array}\right.
	\end{split}
\end{equation}

In Eq. \ref{equ:goal}, $\mathbf{q}_{\text{grasp}}^i$ is the grasp configuration of $\mathcal{R}^i$.
$\mathbf{p}(t)$ defines the pose of object $O$ at time $t$.
$\mathbf{T}(\mathbf{e}_i,\mathbf{e}_j,t)$ defines the relative pose between end-effectors $\mathbf{e}_i$ and $\mathbf{e}_j$, and we use a zero first derivative to indicate that it is a constant matrix, which is also called closed-chain constraints \cite{liu2025planning}.
$\mathbf{G}$ encapsulates the manifold of task-relevant constraints that the object must satisfy throughout the manipulation sequence, including but not limited to obstacle avoidance and force-closure maintenance.
We divide collaborative manipulation into two stages: in the first stage, $\mathcal{R}^i$ moves from its initial pose to $\mathbf{q}_{\text{grasp}}^i$ before $t_1$; then, the collaborative manipulation process begins with constraints $\dot{\mathbf{T}}(\mathbf{e}_i,\mathbf{e}_j,t)=\mathbf{0}$.

We solve this problem in two steps.
The first step is collaborative grasp pose generation, including local grasp pose generation and collaborative selection. 
LoGNet is employed to generate locally stable grasp poses, while the VLM is used to generate final collaborative grasp poses based on object properties and the number of robots. 
Next, a motion planning module is used to generate trajectories.
The overall framework of this work is presented in Fig. \ref{fig:total}.

\section{VLM-Based Inference of Motion Constraints}

Each robot is equipped with a camera. 
Before the manipulation task, initial observations of the environment are required. 
We assume a set of RGB-D images $\{s_i\}$ along with their corresponding camera pose matrices $\{\mathbf{p}_{\text{cam},i}\}$. 
The object $O$ and the structure of the environment are embedded within this information.
Using the acquired RGB images, we employ SEEM\cite{zou2023segment} to detect the precise point clouds of the object. 
SEEM is an open-source foundation model that is capable of performing segmentation with both language and visual inputs. 
We first extract a text description of the target object (e.g., ``table'') based on the given language instruction. 
Then, the text description along with all RGB images is fed into SEEM, which returns a corresponding mask for the object in each image. 
The object point cloud extracted from image $s_i$ is denoted as $P^i_\text{obj}$. 
By transforming all $P^i_\text{obj}$ into the unified world frame, the fused object point cloud $P_\text{obj}$ can be obtained.

Then, we use the VLM to infer object-level constraints during manipulation, including end-pose constraints and motion constraints. 
End-pose constraints refer to the target states that the manipulated objects must achieve, such as requiring the $z$-axis to point upward. 
Motion constraints refer to poses or geometric relations that must be maintained while the objects are being operated.
For example, when moving a table with items on it, the table must be kept horizontal.
Similar to \cite{huang2024copa}, a local coordinate frame of $P_\text{obj}$ can be built and projected onto an image $s_i$.
Then, VLMs can be used to describe the required motion constraints $\mathbf{G}(\mathbf{p}(t))$ based on the local coordinate frame and the world frame. 
Based on the generated constraints, the collaborative grasping and motion planning algorithms can produce appropriate trajectories that satisfy them.

\section{Collaborative Grasp Poses Generation}

\label{sec:grasp}

For SCM tasks, the problem can generally be decoupled into two components. 
The first part involves generating appropriate grasp poses for each robot, while the second part focuses on planning suitable trajectories to reach these grasp poses and executing the manipulation task while satisfying end-effector constraints.
In this section, we study how to generate feasible grasp poses.
In Section \ref{sec:plan}, we will focus on the planning part.

When humans perform collaborative transportation, they typically separate grasping into two parts: \textbf{global coordination} and \textbf{local pose generation}.
In the first stage, humans initially determine rough grasping areas among individuals based on the distribution of feasible grasp sites on the object, thereby reducing the force and torque allocated to each participant. 
In the second stage, each individual focuses on how to select a stable grasp pose within the vicinity of their designated area.
Inspired by this, we also divide collaborative grasping into coordination and local pose generation. 

In this section, we first introduce how to generate local grasp poses, which can be used to filter feasible grasp locations on a large object to generate potential configurations for multi-robot coordination. Then, we utilize a VLM to select an appropriate distribution of these grasp locations among the robots.
% In this section, leveraging the ShapeNet \cite{chang2015shapenet} and DA$^2$ \cite{zhai20222} dataset, we construct a dataset and train LoGNet for local grasp pose generation of large objects. 
% Then, we utilize a VLM to enable grasp pose allocation among multiple robots.

\subsection{Local Grasp Pose Generation}

Existing 6-DoF grasp networks\cite{fang2023anygrasp} are tailored for tabletop objects, leading to severe scale mismatch when they are applied to large targets. 
Directly downsampling a large object's point cloud loses critical local geometric details (e.g., curvature and edges) required for stable grasping. 
To address this, our approach leverages a localized point cloud cropping technique tailored to the physical scale of the gripper. 
This enables our method to generalize across objects of various sizes by effectively decoupling global object size from local grasp synthesis, ensuring high-fidelity feature extraction.
In this work, LoGNet in trained to predict poses near a reference point $\mathbf{p}\in \mathbb{R}^3$, conditioned on the local point cloud $P_i$.

\subsubsection{Dataset Generation}

ShapeNet \cite{chang2015shapenet} is a dataset containing various types of 3D models, including different types of furniture. 
The DA$^2$ \cite{zhai20222} dataset contains diverse grasp poses for large objects from ShapeNet. 

For a given object $o_i$ in the DA$^2$ dataset, we denote the annotated grasp poses in the object frame as $\mathbb{G}^{o_i}$.
For each grasp pose, we also obtain two finger contact points, $\mathbf{p}^1_{\text{con}}(\mathbb{G}^{o_i})$ and $\mathbf{p}^2_{\text{con}}(\mathbb{G}^{o_i})$.
A randomized virtual camera is simulated to observe $o_i$ from a viewpoint that contains the object, and its pose is represented by $\textbf{cam}_j$. 
An RGB-D image is captured using the simulated camera, representing a local observation of $o_i$. 
The corresponding point cloud of this RGB-D image in the object frame is $P^{o_i}_{\textbf{cam}_j}$.
For all grasp poses $\mathbb{G}^{o_i}$, we retain those are observed in the current view by computing their distance to $P^{o_i}_{\textbf{cam}_j}$.
The remaining poses can be denoted as $\mathbb{G}^{o_i}_{\textbf{cam}_j}$.

In addition, a reference point $\mathbf{p}$ is required for LoGNet to indicate where to generate this local grasp pose. 
Based on $\mathbf{p}^1_{\text{con}}(\mathbb{G}^{o_i})$ and $\mathbf{p}^2_{\text{con}}(\mathbb{G}^{o_i})$, we randomly select one of them and add a random offset to avoid potential dataset leakage.
The resulting grasp point captured by $\textbf{cam}_j$ is denoted by $\mathbf{p}_{\textbf{cam}_j}(\mathbb{G}^{o_i})$.
The local point cloud around this grasp point, $P_{\text{local}}(\mathbb{G}^{o_i}_{\textbf{cam}_j})$, is defined as
\begin{equation}
	\label{equ:range}
	P_{\text{local}}(\mathbb{G}^{o_i}_{\textbf{cam}_j}) =
	\{\mathbf{x}\in P^{o_i}_{\textbf{cam}_j}\mid
	\|\mathbf{x} - \mathbf{p}_{\textbf{cam}_j}(\mathbb{G}^{o_i})\|_{\infty} < r\}.
\end{equation}
where $r$ is the crop radius determined by the gripper scale (we set $r$ to 1).
Then, $P_{\text{local}}(\mathbb{G}^{o_i}_{\textbf{cam}_j})$ is translated to center the reference point, indicating where the grasp pose should be generated.
Finally, the obtained point cloud is transformed into the camera frame and downsampled to 2048 points to obtain $\hat{P}_{\text{local}}(\mathbb{G}^{o_i}_{\textbf{cam}_j})$, which is the final input of LoGNet.

\begin{figure}[!t]
	\centering   
	\includegraphics[width=3.3in]{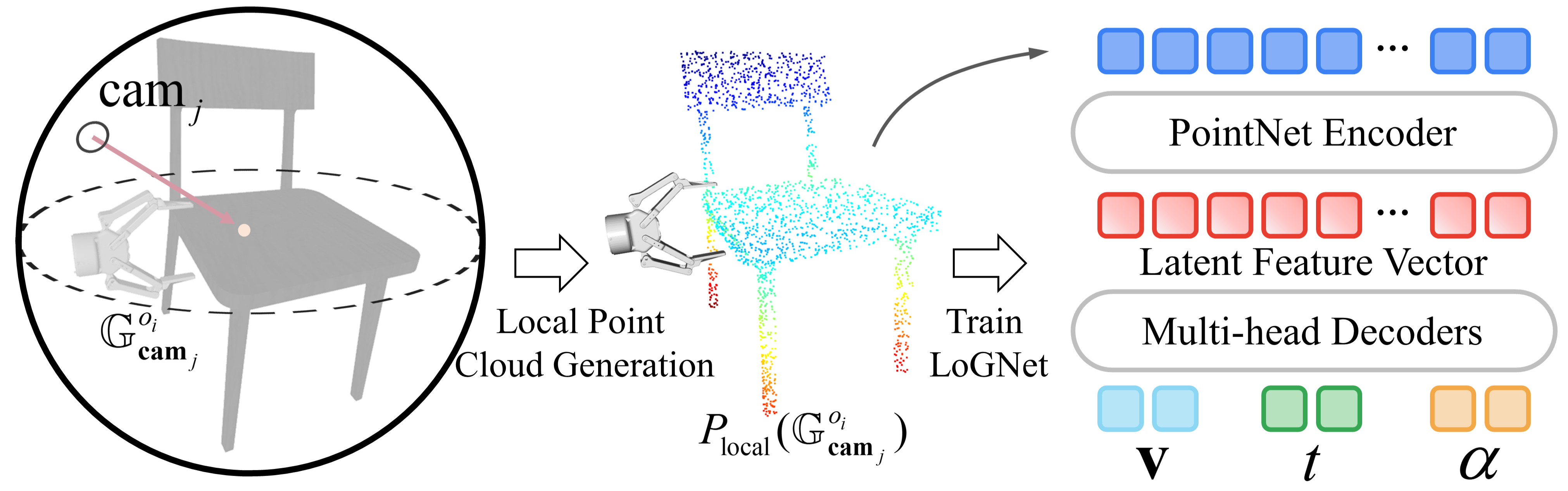}
	\caption{
		Illustration of local grasp pose generation.
		Local point clouds of multiple objects are captured using cameras.
		Then, LoGNet is trained for local grasp pose generation.
	}
	\label{fig:local_pc}
\end{figure}

\subsubsection{Network Architecture}
%backbone, input and output of the NN
Considering the efficiency of training, we adopt PointNet\cite{qi2017pointnet} as the backbone network for LoGNet. 
The backbone takes a $2048 \times 3$ point cloud as input and outputs a 1024-dimensional feature vector. 
The target output is a grasp pose $\mathbb{G}$ representing a transformation matrix in $\mathbb{R}^{4 \times 4}$.
$\mathbb{G}$ can be decomposed into a translation component $\mathbf{t}$ and a rotation component $\mathbf{R}$. 
$\mathbf{t}$ is relatively easy to learn. 
However, $\mathbf{R}$, having only three degrees of freedom, is inherently constrained. 
Following the implementation used in AnyGrasp\cite{fang2023anygrasp}, we decompose the rotation matrix into an approach vector $\mathbf{v}$ (i.e., the direction of the gripper) and an in-plane rotation $\alpha$ along the approach vector (i.e., the orientation of the gripper's opening direction).
Using MLPs, the 1024-dimensional feature vector is mapped through three operation networks to predict $\mathbf{t}$, $\mathbf{v}$, and $\alpha$, respectively.

% It is important to note that we do not regress grasp scores and widths. 
% As for the former, since this work focuses on collaborative grasping, local grasp quality metrics such as force closure are insufficient to reflect the overall grasp performance. 
% In experiments, we instead evaluate the results based on global metrics such as overall force closure. 
% For width, although it can be easily integrated into the framework, we choose not to output it because the gripper used in experiments does not require specifying a grasp width.
\subsubsection{Training Detail}
%Loss function
Our loss function is 
\begin{equation}
	\label{equ:loss}
	\mathcal{L}=\sum_{(o_i,\textbf{cam}_j,\mathbb{G})\in \mathcal{D}} \lambda_1\|\hat{\mathbf{v}}-\mathbf{v}\|_2 + \lambda_2\|\hat{\mathbf{t}}-\mathbf{t}\|_2 + \lambda_3\sin^2(\hat{\alpha}-\alpha),
\end{equation}
where $\mathcal{D}$ denotes the generated local grasp dataset, $\hat{\mathbf{v}}$, $\hat{\mathbf{t}}$, and $\hat{\alpha}$ are the predicted values, and $\mathbf{v}$, $\mathbf{t}$, and $\alpha$ are the corresponding ground-truth values.
$\lambda$ represents the weights of different components in $\mathcal{L}$, and we set all of them to 1.
We collect a total of 752 indoor large object models. 
For each object, we generate 10 uniformly distributed random rotations in $\mathrm{SO}(3)$ and apply them to the object, which has a similar effect to capturing the object from different camera poses. 
In total, we obtain 88,976 local point clouds and corresponding local grasp poses.
To avoid dataset leakage, we perform data splitting at the object level, using 70\% for training and the remaining 30\% for testing.
The framework for local grasp pose generation can be found in Fig. \ref{fig:local_pc}.

\subsection{Collaboration between Different Robots}

In this section, we introduce how VLMs can be leveraged to determine the grasp point $\mathbf{p}$ for each robot.

\subsubsection{Grasp Points Projection}
Given the object point cloud $P_\text{obj}$, we first downsample it to obtain a set of candidate grasp points. 
We then utilize LoGNet to generate grasp poses for these candidates. 
Based on $P_\text{obj}$, collision checking is performed between the gripper model under these poses and $P_\text{obj}$ to filter out invalid configurations. 
The remaining points, where collision-free grasp poses can be successfully generated, are denoted as $P_{\text{CF}}$.
To prevent label overlap, which could confuse the VLM regarding individual grasp locations, we select potential grasp points exclusively from a single image's point cloud $s_i$. 
These candidate grasp locations are then projected onto a single image. 
This allows us to fully exploit the VLM's spatial understanding to determine the appropriate position for global coordination.

The collision-free point cloud in image $s_i$ can be denoted as $P^i_{\text{CF}}=P_{\text{CF}} \cap P^i_\text{obj}$. 
Then, we apply k-Medoids clustering to $P^i_{\text{CF}}$ to obtain $K$ ($K>N$) potential grasp pose candidates. 
The use of clustering helps ensure that the generated grasp poses are well distributed across the object, which promotes balanced grasp poses.
Finally, the $K$ generated grasp poses are projected onto the image $s_i$ to obtain $\hat{s}_i$, and each grasp pose is annotated with a number from 1 to $K$. 
An example of an annotated image can be found in Fig. \ref{fig:total}. 
A VLM can be used to select the grasp poses based on $\hat{s}_i$.

\subsubsection{Grasp Points Selection}
The primary challenge in this part is enabling the VLM to ground the input labeled image $\hat{s}_i$ in the grasping task. Due to limited spatial understanding, VLMs often struggle to associate textual labels with their exact physical locations on an object, leading to severe hallucinations during grasp selection. Furthermore, they exhibit an inherent deficiency in understanding the implicit mapping between the labeled positions in a 2D image and the eventual execution success rate of the physical collaborative manipulation. We aim to mitigate these limitations in this section.

We use an in-context learning approach to achieve this goal.
Our prompt design consists of three main components. 
First, we define the specific metrics that characterize a successful collaboration. 
Second, a Chain-of-Thought (CoT) \cite{wei2022chain} strategy is employed to force the VLM to reason about the underlying relationships between the text instructions and the physical objects. 
Finally, we provide structured examples to guide the VLM in adhering to the desired output format.

Inspired by human decision-making during grasping, we introduce three metrics for VLM to follow.
\begin{itemize}[leftmargin=*]
	\item Collaboration between all robots. The VLM should ensure that the object’s center of mass remains within the convex hull of the grasp points to reduce gravitational torque. In addition, the grasp poses should be well dispersed to avoid inter-robot collisions during motion.
	\item The angle between the gripper and the external force, e.g., gravity. A two-finger gripper can achieve a more stable grasp with less force if the line connecting its fingers is parallel to the direction of the external force. 
	\item The local stability of the grasp point. The object shape and surface properties, such as friction, are considered to reduce grasp slipping.
\end{itemize}

Inspired by CoT, we first prompt the VLM to identify the precise location of each numerical label within the object, requiring it to explicitly analyze these positions in its output. 
We find that this approach effectively prevents the VLM from hallucinating during the grasp point selection.
Furthermore, to ensure the VLM strictly adheres to evaluation metrics and CoT, we provide some output examples for guidance.

Ultimately, the VLM outputs a set of suitable numbers to represent those grasp points. 
By integrating this with LoGNet, we can generate an appropriate set of grasp poses for cooperative manipulation tasks.

\section{Collaborative Motion Planning}

\label{sec:plan}
Based on the end-effector grasp poses for all robots, initial collision-free poses in configuration space $\mathbf{q}_{\text{grasp}}^i$ for each robot can be obtained through inverse kinematics.

Collaborative motion planning is divided into two parts: 1) moving the assigned robots from their initial poses to the grasping poses $\mathbf{q}_{\text{grasp}}^i$; and 2) planning an object trajectory that satisfies all constraints, while each robot plans its own trajectory to follow the object trajectory.

For the first stage, this is a typical multi-robot path planning problem. 
%The core difficulty lies in resolving path conflicts. 
%However, since this is not the main research focus of this work, 
We abstract the mobile manipulator's geometry into a series of bounding spheres to enable rapid collision detection and efficient path planning within the environment.
We employ RRT* in the configuration space to obtain collision-free paths for the robots sequentially. 
For the second part, given the object end pose $\mathbf{p}(t_e)=\mathbf{p}_{\text{end}}$ and motion constraints $\mathbf{G}(\mathbf{p}(t))$, we use RRT* to generate an object trajectory firstly.
Once the object motion and grasp points are given, the end-effector trajectories can be obtained.
Then, a reactive control-based approach \cite{haviland2021neo} is applied to generate the motion of each robot while considering collisions.
Due to that object planning does not explicitly consider robot-body collisions during robot motion planning, robots may fail to find feasible trajectories under a given object trajectory. 
In such cases, these object poses are treated as collision-prone poses, and the object trajectory is replanned.
If feasible trajectories cannot be found after 3 replanning attempts, the current collaborative grasp poses are regarded as infeasible and ignored.
Then, grasp poses are regenerated for motion planning.
% A more detailed description of our motion planning algorithm will be presented in appendix on the project page.

Based on the path planning module described above, we can generate robot motion trajectories to accomplish the given task $\mathcal{T}$. 
For the actual trajectory execution, we apply PID control to enable each robot to accurately track the specified trajectory.

\section{Experiments}
\begin{figure*}[!t]
	\centering   
	\includegraphics[width=7in]{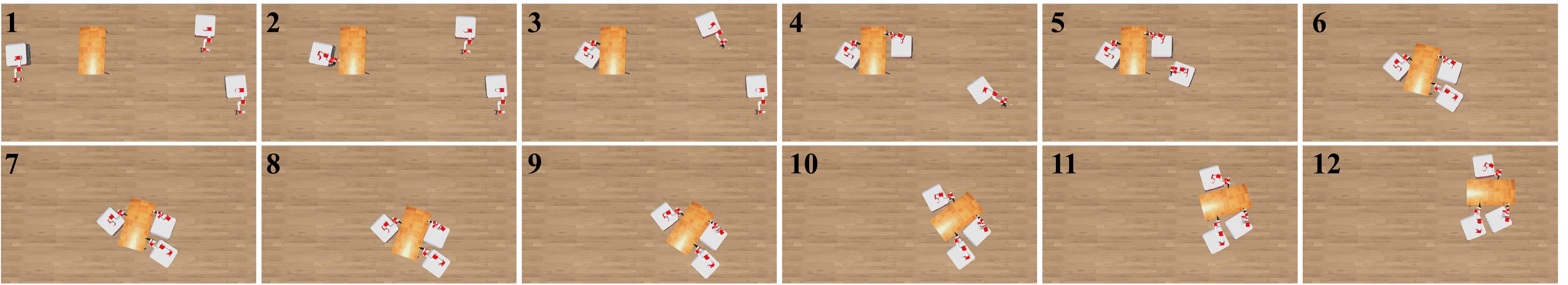}
	\caption{
		An example of using 3 robots to move a table.
		Frames 1--6: collaborative grasping is achieved. Frames 7--12: SCM is performed.
	}
	\label{fig:case_study}
\end{figure*}

Experiments are conducted in both simulation and the real world.
The mobile platforms are omnidirectional wheeled bases equipped with JAKA Zu7 manipulators.
The end-effector of each manipulator is a Robotiq 2F-85 gripper. 
The system employs a centralized communication architecture, where images and states are transmitted to a host computer. 
The host then plans the trajectories for all robots to execute the task. 
We assume that the initial positions are known and that each robot can maintain accurate self-localization during motion. 
In real-world experiments, the relative initial positions are pre-defined, and self-localization is performed via wheel odometry.
To evaluate our system, our experiments are designed to address the following three key research questions:
\begin{itemize}[leftmargin=*]
	\item \textbf{Grasp Feasibility:} Can the local grasp module consistently generate stable and high-quality grasp poses?
	\item \textbf{Framework Effectiveness:} What is the overall success rate of the integrated framework when combining both grasp generation and planning?
	\item \textbf{Real-world Deployability:} Can the proposed system be successfully deployed on physical robot platforms?
\end{itemize}
We provide comprehensive comparisons, ablation studies, and failure case analyses to answer these questions, alongside a simulation case study illustrated in Fig. \ref{fig:case_study}.

\subsection{Grasp Poses Evaluation}

\subsubsection{Comparison of Generated Grasp Poses}
%local grasp pose generation, compare with graspnet
\begin{figure}[!ht]
	\centering   
	\includegraphics[width=3.3in]{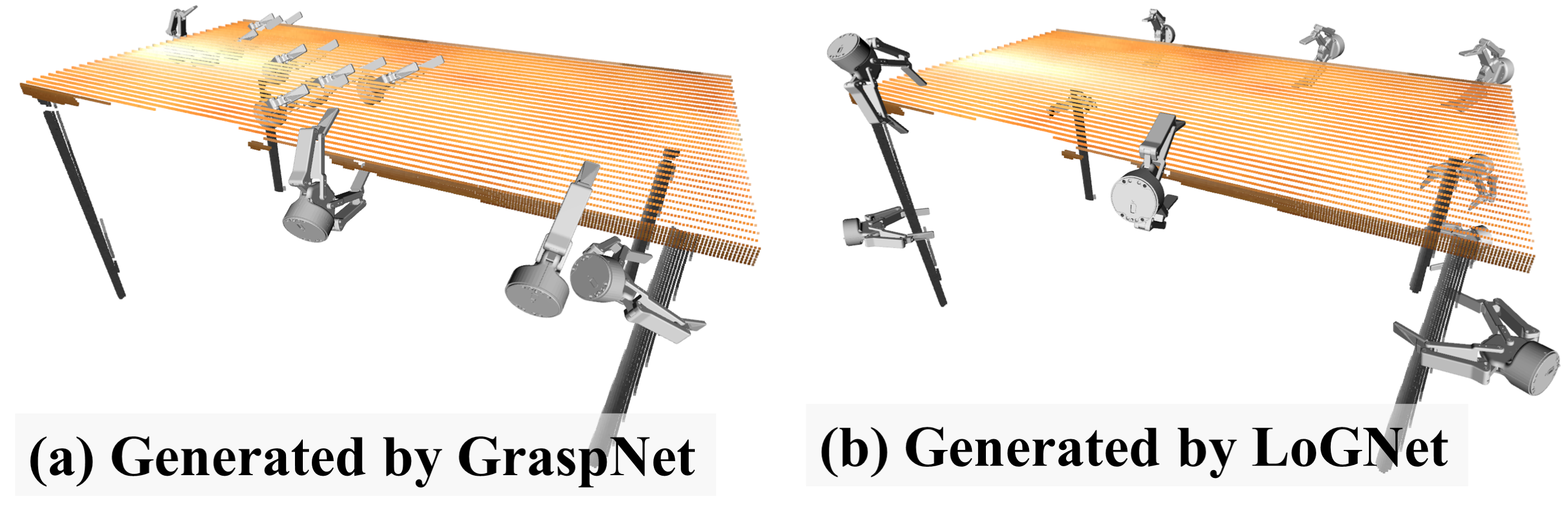}
	\caption{
		Comparison of GraspNet \cite{fang2020graspnet} and LoGNet. LoGNet is capable of generating high quality grasp poses.}
	\label{fig:grasp_compare}
\end{figure}

To verify the effectiveness of the proposed LoGNet, we compare it with a representative method, GraspNet \cite{fang2020graspnet}. 
The result is presented in Fig. \ref{fig:grasp_compare}.
It can be observed that GraspNet has poor performance in generating grasp poses for large objects, which can be attributed to the network's insufficient perception of local structures in large object point cloud.
Moreover, the grasp poses generated by GraspNet can collide with large objects. 
In contrast, LoGNet is capable of synthesizing high-quality, viable grasp poses even when conditioned on a single input image, which inherently yields incomplete and partial target point clouds.
This provides a solid foundation for subsequent grasp selection and collaborative manipulation.

\subsubsection{Evaluation of Collaborative Grasp Poses Generation}
\begin{table}[!ht]
	\centering
	\caption{Comparison of grasp poses selection under different VLMs. $\omega$ ($\downarrow$), MSV ($\uparrow$), $f_\text{max}$ ($\downarrow$).}
	\renewcommand{\arraystretch}{1.2}
	\small
	\begin{tabular}{|l|ccc|ccc|} % 更新为 7 列 (1列标签 + 2个模型 * 3个指标)
		\hline
		\multirow{2}{*}{Metrics} & \multicolumn{3}{c|}{GPT 4.1} & \multicolumn{3}{c|}{Gemini 2.5 Pro} \\
		\cline{2-7} % 更新横线跨度
		& $\omega$ & MSV & $f_{\text{max}}$ & $\omega$ & MSV & $f_{\text{max}}$ \\
		\hline
		\small TABLE-2 & 0.31 & 0.031 & 16.5 & 0.06 & 0.030 & 7.30 \\
		\small TABLE-3 & 0.02 & 2.767 & 0.54 & 0.02 & 0.944 & 0.48 \\
		\small Tilted TABLE & 0.03 & 1.416 & 1.63 & 0.30 & 2.693 & 2.75 \\
		CHAIR & 0.10 & 0.033 & 1.08 & 2.07 & 0.043 & 1.52 \\
		\hline
	\end{tabular}
	\label{tab:compare_vlm}
\end{table}
% \begin{table*}[!ht]
	% 	\centering
	% 	\caption{Comparison of grasp poses selection under different VLMs. GS ($\uparrow$), $\omega$ ($\downarrow$), MSV ($\uparrow$), $f_\text{max}$ ($\downarrow$).}
	% 	\renewcommand{\arraystretch}{1.2}
	% 	\small
	% 	\begin{tabular}{|l|cccc|cccc|cccc|cccc|} % Updated column definition
		% 		\hline
		% 		\multirow{2}{*}{Metrics} & \multicolumn{4}{c|}{QWEN-VL-MAX} & \multicolumn{4}{c|}{DeepSeek-R1} & \multicolumn{4}{c|}{GPT 4.1} & \multicolumn{4}{c|}{Gemini 2.5 Pro} \\
		% 		\cline{2-17} % Updated cline to span 16 columns (4 models * 4 sub-columns)
		% 		& GS & $\omega$ & MSV & $f_{\text{max}}$& GS & $\omega$ & MSV & $f_{\text{max}}$ & GS & $\omega$ & MSV & $f_{\text{max}}$ & GS & $\omega$ & MSV & $f_{\text{max}}$ \\
		% 		\hline
		% 		\small TABLE-2 & $\text{\ding{55}}$ & 3.09 & 0.023 & / & $\text{\ding{55}}$ & 1.73 & 0.023 & / & \ding{51} & 0.31 & 0.031 & 16.5 & \ding{51} & 0.06 & 0.030 & 7.30 \\
		% 		\small TABLE-3 & $\text{\ding{55}}$ & 2.13 & 2.379 & / & $\text{\ding{55}}$ & 1.48 & 1.327 & 1.22 & \ding{51} & 0.02 & 2.767 & 0.54 & \ding{51} & 0.02& 0.944 & 0.48 \\
		% 		\small Tilted TABLE & $\text{\ding{55}}$ & 0.08 & 1.163 & 7.77 & $\text{\ding{55}}$ & 0.98 & 2.626 & / & \ding{51} & 0.03 & 1.416 & 1.63 & \ding{51} & 0.30 & 2.693 & 2.75 \\
		% 		CHAIR & $\text{\ding{55}}$ & 0.41 & 0.025 & 1.55 & $\text{\ding{55}}$ & 0.10 & 0.029 & 2.04 & \ding{51} & 0.10 & 0.033 & 1.08 & \ding{51} & 2.07 & 0.043 & 1.52 \\
		
		% 		\hline
		% 	\end{tabular}
	% 	\label{tab:compare_vlm}
	% \end{table*}

%In this section, we evaluate the collaborative grasp poses generation module.
The most critical and difficult part of SCM is collaborative grasping.
We validate that our proposed method works under varying object types and robot numbers.
Our method is tested using two different VLMs: GPT 4.1, and Gemini 2.5 Pro.

We employ quantitative metrics to evaluate whether the VLM is capable of selecting appropriate collaborative grasp poses.
Three metrics are used to evaluate the result.
Similar to \cite{zhai20222,liu2021synthesizing}, $G$ is used to represent the grasp matrix.
A relaxation of the force closure metric is used \cite{liu2021synthesizing}
\begin{equation}
	\omega=\|G\mathbf{c}\|_2 = \|G\textbf{f}_t\|_2/\|\textbf{f}_n\|_2,
\end{equation}
where $\mathbf{c}$ is the set of friction cone axes, and $\textbf{f}_n$ and $\textbf{f}_t$ are the normal and tangential components of the contact force.
A smaller $\omega$ means force closure is more likely to be achieved.
The minimum singular value (MSV) of $G$ \cite{zhai20222} is used to evaluate the grasp quality.
A larger MSV will lead to a more symmetrical distribution of grasp points around the centroid of the object.
Further, we evaluate the results by calculating the contact forces required to perform the task under the influence of gravity.
We calculate a feasible solution for all contact forces, which can be written as
\begin{equation}
	\label{equ:cal_f}
	\begin{aligned}
		\min_\mathbf{f} & \ \mathbf{f}^T \mathbf{f}\\
		\text{s.t.} &\ GG^\top \succeq \epsilon I_{6 \times 6}, \ G\mathbf{f} + \mathbf{f}_{\text{ext}} = \mathbf{0},\\
		&\ \mathbf{f}_i^T \mathbf{c}_i > \|\mathbf{f}_i\|_2/\sqrt{\mu^2 + 1}, \\
	\end{aligned}
\end{equation}
where $\mathbf{f}$ is the combination of all contact forces.
$\mathbf{f}_{\text{ext}}$ denotes the external wrench.
We consider gravity $\textbf{f}_g$ as its dominant component.
In Eq. \ref{equ:cal_f}, there are three constraints: the first means $G$ is full rank; the second enforces contact-force equilibrium with the external wrench; and the third requires each contact force to lie within its friction cone.
Since there can be multiple contact-force solutions $\mathbf{f}$ satisfying these constraints, we obtain a particular feasible solution $\mathbf{f}^*$ by minimizing the L2 norm of $\mathbf{f}$.
We define the maximum ratio between each optimal contact force $\mathbf{f}^*_i$ and gravity $\mathbf{f}_g$ as $f_{\text{max}}$
\begin{equation}
	f_{\text{max}} = \max_i \|\textbf{f}^*_i\|_2/\|\textbf{f}_g\|_2.
\end{equation}
A smaller $f_{\text{max}}$ indicates that the task can be achieved more easily based on the selected grasp points. 
This metric is effective for evaluating both the gripper-force angle and the collaboration among all robots.

In Table \ref{tab:compare_vlm}, four different scenarios are considered.
We primarily consider the generalization ability of the proposed grasping module under three different scenarios:
1) varying numbers of robots;
2) different types of manipulation tasks;
3) different categories of manipulated objects.

First, for TABLE-3, 3 robots are used, and the number of candidate grasp points $K$ is set to 15. 
In the other experiments, 2 robots and 10 candidate grasp points are used. 
In the Tilted TABLE task, the objective is to upright a tilted table, which primarily involves a rotation motion. 
The other three tasks mainly involve translational motions. 
Finally, we select two common large indoor objects, a table and a chair, as the target objects for manipulation.

Detailed results are presented in Table \ref{tab:compare_vlm}.
First, regarding the VLM's ability to ground visual inputs to grasp points, both GPT-4.1 and Gemini 2.5 Pro accurately identify the labeled grasp points on the objects. Notably, even in uncommon scenarios such as the tilted table scene, both models consistently recognize the precise positions of various grasp points, demonstrating strong robustness. Consequently, this high-fidelity spatial grounding directly translates to superior grasp quality, enabling both VLMs to successfully accomplish the grasping tasks across all four evaluated scenarios.

For the table-moving task, increasing the number of robots to three leads to a significant reduction of $f_{\text{max}}$, indicating that the maximum grasping force required by each robot is greatly reduced. 
This demonstrates the advantage of multi-robot collaboration. 
Comparing the rotation task (Tilted TABLE) and the translation task (TABLE-2), the proposed method effectively solves the grasping challenges in both scenarios. 
%Since various manipulation tasks can be viewed as combinations of rotation and translation, this suggests that our method can be extended to a wide range of tasks, demonstrating its generalization capability. 
Finally, the method also shows strong generalization across different object types, specifically for both TABLE and CHAIR.

%Through this experiment, we validate the robustness of the grasping module, thereby laying a solid foundation for the overall framework.

\subsubsection{Ablation Study on the Prompt Design}
%This work addresses the problem of grasp point selection using an in-context learning approach. 
%The prompt design plays an important role and incorporates three distinct modules. 
%In this section, we aim to demonstrate that a properly designed prompt enhances the spatial understanding capabilities of the VLM, thereby enabling successful collaborative grasp pose generation.
%First, our prompt provides evaluation metrics for grasp point selection. 
%Second, it uses CoT to ground the labeled images with the grasping task, analyzing the position and properties of each grasp point. 
%Finally, by providing examples, it ensures that the VLM's output adheres to the specified CoT format.

In this section, we demonstrate that a properly designed prompt improves the VLM's spatial understanding for collaborative grasp pose generation. The prompt defines grasp evaluation metrics, uses CoT to analyze grasp point properties from labeled images, and provides examples to enforce the desired output format.
%In this section, an ablation study is conducted to validate the effectiveness of prompt design.

\begin{table} 
	\caption{Ablation study on the prompt design. Metrics in the table: mean $\pm$ std. of $f_{\text{max}}$ (success count). A: evaluation metrics; B: CoT; C: output example. }
	\centering 
	\renewcommand{\arraystretch}{1.2}
	\begin{tabular}{c c c}
		\hline
		& Gemini 2.5 Pro & GPT 4.1 \\
		\hline
		w/o ABC & $5.40\pm 3.79\ (2)$ & $26.06\pm 33.01\ (2)$ \\
		w/o BC & $20.36\pm 25.15\ (3)$ & $32.94\pm\ 23.27\ (2)$ \\
		w/o B & $20.36\pm 25.15\ (3)$ & $31.83\pm\ 24.84\ (2)$ \\
		w/o C & $4.71\pm 1.96\ (3)$ & $26.72\pm 19.67\ (3)$ \\
		Ours & $\mathbf{4.57\pm 2.20\ (3)}$ & $23.71\pm 22.95\ (3)$ \\
		\hline
	\end{tabular}
	\label{tab:abl_grasp}
\end{table}

We evaluate two VLMs with label-to-image grounding capability. 
Images of the table from three different perspectives are used for evaluation.
Results can be found in Table \ref{tab:abl_grasp}.
Without providing evaluation metrics (w/o ABC), both VLMs produce sub-optimal grasp selections, leading to failures in completing some grasping tasks. 
However, by introducing these metrics, we observe a significant improvement for reasoning models. 
Gemini 2.5 Pro can achieve reasonable analysis and obtain favorable results, thereby increasing the success count. 
In contrast, for a non-reasoning model such as GPT 4.1, simply introducing evaluation metrics is insufficient.
If we provide an output example without CoT, the introduced example has little influence on the final result.
This is because these examples contain only the output format and do not improve the selection strategy.
When evaluation metrics and CoT are provided, all VLMs can select reasonable grasp points.
When all three modules are used, both VLMs show the best results, indicating the effectiveness of the proposed method.

\subsection{Overall Success Rate}

In this section, we evaluate the overall success rate of this system.
Five different tasks are evaluated.
The detailed results are presented in Table \ref{tab:success_rate} where we use Tbl. and Chr. to represent table and chair respectively.
In Tbl. and Chr. tasks, translation and horizontal rotation are applied to these two objects.
In Flip Chr., a vertical rotation, like flipping, is applied, which verifies the generalization capability of uncommon tasks.
In the Align task, we move a chair near a table.
Except for Tbl.-3, where 3 robots are used, other tasks use 2 robots.
For each task, 5 different images are used to represent the object being manipulated. 
In addition, the initial poses of the robots are \textbf{randomly generated} for each task.

\begin{table} 
	\caption{Success rate on varying tasks.}
	\centering 
	\renewcommand{\arraystretch}{1.2}
	\begin{tabular}{c c c c c c}
		\hline
		SR & \small Tbl. & \small Chr. & \small Flip Chr. & \small Align&  \small Tbl.-3 \\
		\hline
		CollaBot & 8/10 &  7/10 & 6/10 &  9/10 & 6/10\\
		DP\cite{chi2025diffusion} & 0/10 &  0/10 & 0/10 &  0/10& 0/10\\
		% PPO & 0/5 &  0/5 & 0/5 &  0/5& 0/5\\
		\hline
	\end{tabular}
	\label{tab:success_rate}
\end{table}

We use Gemini 2.5 Pro in the remaining experiments.
The overall success rate of the proposed system is 72\%.
For two-robot translation tasks, including table moving, chair moving, and alignment, the success rate reaches 80\%, higher than that of the flipping task. 
There are two main reasons for this. 
First, in translation tasks, the initial state of the object is generally more common. 
Thus, foundation models such as SEEM and VLMs are more effective at object recognition and grasp point selection, leading to more accurate object point clouds and higher-quality grasp poses.
Second, in flipping tasks, the significant vertical displacement can cause previously accessible grasping positions to become unreachable because they exceed the workspace, thereby increasing the likelihood of failure.
By contrast, translation motions involve only horizontal object movement and can be more easily tracked by the robot base, resulting in a higher success rate.

Then, we investigate the scalability of CollaBot by increasing the team size. 
For the challenging task of moving a table with three robots, a success rate of 60\% is achieved. 
Although the increased coordination complexity and tight closed-chain constraints lead to a performance drop compared to the two-robot scenarios, this result demonstrates the viability of our framework beyond a dual-agent setup, showcasing its potential to scale to larger robot teams.
The main reason for failure is that as the number of robots increases, the likelihood of collisions between robots under given grasp poses rises.
%Therefore, collision-aware grasp selection and trajectory planning become increasingly important as the robot team grows.

We compare our approach with Diffusion Policy\cite{chi2025diffusion}, a representative imitation learning baseline.
The images and proprioceptive states from all robots in the scene are provided as inputs to directly predict the control actions for each robot. 
We utilize CollaBot as the expert policy to collect 50 demonstrations for each task to train the network, which is subsequently evaluated in simulation. 
The final success rates are reported in Table \ref{tab:success_rate}. 
As observed, the imitation learning-based baseline fails to accomplish the tasks.

%Regarding the failure causes of CollaBot, 50\% are due to failures to generate collaborative grasp poses.
%For example, the generated local grasp poses might cause collisions, or the collaborative grasp poses selected by the VLMs might cause inter-robot collisions. 
%33.3\% of failures stem from planning issues, such as path conflicts. 
%The remaining 16.7\% are caused by SEEM failing to detect the object point cloud, which is particularly common in flipping tasks.

A failure case analysis is further conducted.
3 cases are caused by unstable local grasp pose generation, where the generated grasp poses cannot maintain stable contact during manipulation. 
4 failures originate from collaborative grasp pose selection. Although the selected grasp points are physically reasonable, the corresponding multi-robot inverse kinematics (IK) problems become infeasible due to limited workspace overlap between robots.
6 cases stem from the motion planning stage. 
These failures are mainly caused by robot-environment collisions during collaborative transportation.
Finally, 1 failure is caused by violations of the end-effector closed-chain constraint during execution.
%These results suggest that constrained collaborative motion planning and grasp feasibility remain the primary bottlenecks of large-object SCM tasks.

%We also evaluate the time consumption of different components.
%Take the alignment task as an example, there are five time-consuming steps: VLMs' inference (25.9\,s), local grasp poses generation (96.3\,s), collaborative grasping (21.1\,s), motion planning stage 1 (6.6\,s), and stage 2 (28.0\,s).
%Since our method generates trajectories offline and is able to produce high-quality trajectories within a suitable time, this validates the effectiveness of the proposed framework.

\subsection{Real-world Experiment}

\begin{figure}[!t]
	\centering   
	\includegraphics[width=3.3in]{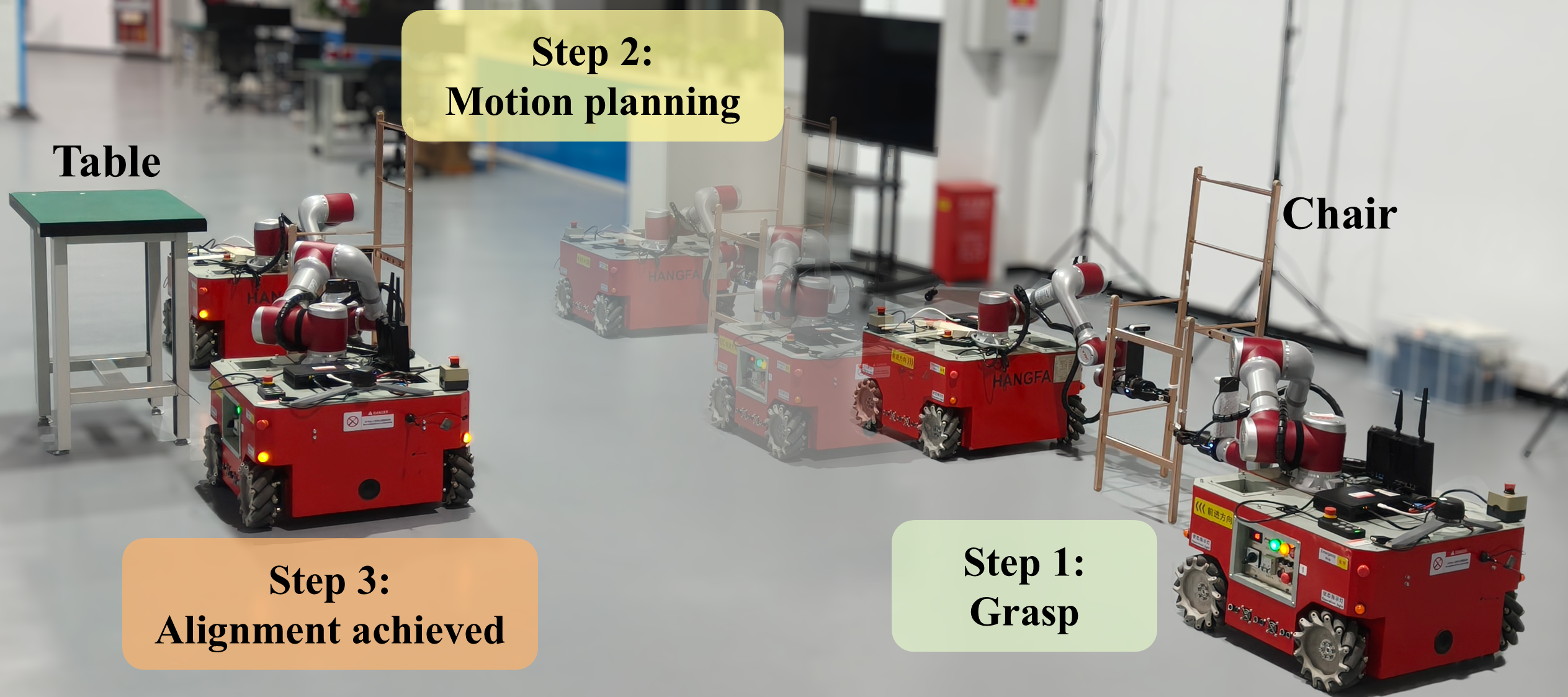}
	\caption{
		Real-world deployment of CollaBot with two robots.}
	\label{fig:rwexp}
\end{figure}

We perform a real-world experiment to demonstrate the feasibility of CollaBot.
Using two robots, we executed a task to move a chair near a table. 
CollaBot successfully generated real-world trajectories, and the robots took 354\,s to execute these trajectories, validating the effectiveness of our proposed method.
The result is presented in Fig. \ref{fig:rwexp}.

\section{Conclusion}
In this work, we introduce CollaBot for SCM tasks using multiple mobile manipulators.
Given a subtask, we use foundation models to detect the target object and infer constraints.
Then, we propose a collaborative grasping module that decomposes grasp generation into local grasp pose generation and collaboration among multiple robots.
Finally, a planning framework is proposed to generate a feasible trajectory.
The overall success rate of this system is 72\%, which demonstrates the effectiveness of this framework and its generalizability across different tasks.

The main limitation of the current work is that it primarily relies on offline planning. 
As a result, it is not applicable to dynamic environments with moving obstacles.
Future work could incorporate a more reactive replanning mechanism to improve robustness for real-world deployment.

\bibliographystyle{ieeetr} 
\bibliography{ref}
\end{document}